\documentclass[10pt,twocolumn,letterpaper]{article}

\usepackage{cvpr}              % To produce the CAMERA-READY version
\definecolor{cvprblue}{rgb}{0.21,0.49,0.74}
\usepackage[pagebackref,breaklinks,colorlinks,allcolors=cvprblue]{hyperref}

\usepackage{bm}
\usepackage{multirow}
\usepackage{colortbl}
\usepackage{xcolor}

\def\paperID{10892} % *** Enter the Paper ID here
\def\confName{CVPR}
\def\confYear{2026}

\title{Cross-Layer Vision Smoothing: Enhancing Visual Understanding via Sustained Focus on Key Objects in Large Vision-Language Models}

\author{
    Jianfei Zhao\textsuperscript{\rm 1,2},
    Feng Zhang\textsuperscript{\rm 1},
    Xin Sun\textsuperscript{\rm 1},
    Chong Feng\textsuperscript{\rm 1},
    Zhixing Tan\textsuperscript{\rm 3}\\
    \textsuperscript{\rm 1}Beijing Institute of Technology\\
    \textsuperscript{\rm 2}Zhongguancun Academy\\
    \textsuperscript{\rm 3}Tsinghua University\\
     {\tt\small zhqingan@bit.edu.cn}
}
\begin{document}
\maketitle
\begin{abstract}
Large Vision-Language Models (LVLMs) can accurately locate key objects in images, yet their attention to these objects tends to be very brief.
Motivated by the hypothesis that sustained focus on key objects can improve LVLMs’ visual capabilities, we propose \textbf{C}ross-\textbf{L}ayer \textbf{V}ision \textbf{S}moothing (CLVS).
The core idea of CLVS is to incorporate a vision memory that smooths the attention distribution across layers.
Specifically, we initialize this vision memory with position-unbiased visual attention in the first layer.
In subsequent layers, the model’s visual attention jointly considers the vision memory from previous layers, while the memory is updated iteratively, thereby maintaining smooth attention on key objects.
Given that visual understanding primarily occurs in the early and middle layers of the model, we use uncertainty as an indicator of completed visual understanding and terminate the smoothing process accordingly.
Experiments on four benchmarks across three LVLMs confirm the effectiveness and generalizability of our method.
CLVS achieves state-of-the-art overall performance across a variety of visual understanding tasks and attains comparable results to the leading approaches on image captioning benchmarks.\footnote{The code is available at: \url{https://github.com/beta-nlp/CLVS}.}

\end{abstract}    
\section{Introduction}
\label{sec:intro}
Large Vision-Language Models (LVLMs)~\cite{llava, qwen25-vl} acquire the capability to comprehend visual information through training on large-scale multimodal data and exhibit strong performance across a wide range of visual understanding tasks~\cite{POPE, AMBER, MME}.
Despite these successes, LVLMs still suffer from issues such as hallucinations~\cite{VCD}, which limit their real-world applications.

Recent studies~\cite{ViCrop, AdaptVis} have revealed a strong correlation between the visual understanding capability of LVLMs and their visual attention. Visual attention is learned through end-to-end training without direct supervision.
While this approach simplifies LVLM training, it also makes it difficult to avoid biases in visual attention~\cite{PAI, VAR, DAC}, such as position bias~\cite{DAC, IMCCD}, visual attention sink~\cite{opera, VAR}, and uncoordinated attention heads~\cite{VHD, AD-HH}. 
Consequently, a series of recent works have sought to enhance the visual understanding capabilities of LVLMs by improving visual attention.
Some works~\cite{PAI, IBD, VAF} propose directly increasing the attention weights on visual tokens.
other works~\cite{AD-HH, VHD} analyze the effect of the multi-head attention mechanism on visual understanding and remove attention heads that negatively impact performance.
Further, several methods~\cite{VAR, TARAC, AdaptVis} adjust the visual attention distribution to achieve finer control over the model’s visual understanding process.
These successes confirm the crucial role of visual attention in the visual understanding of LVLMs.

\begin{figure*}[htbp]
  \centering
  \includegraphics[width=\linewidth]{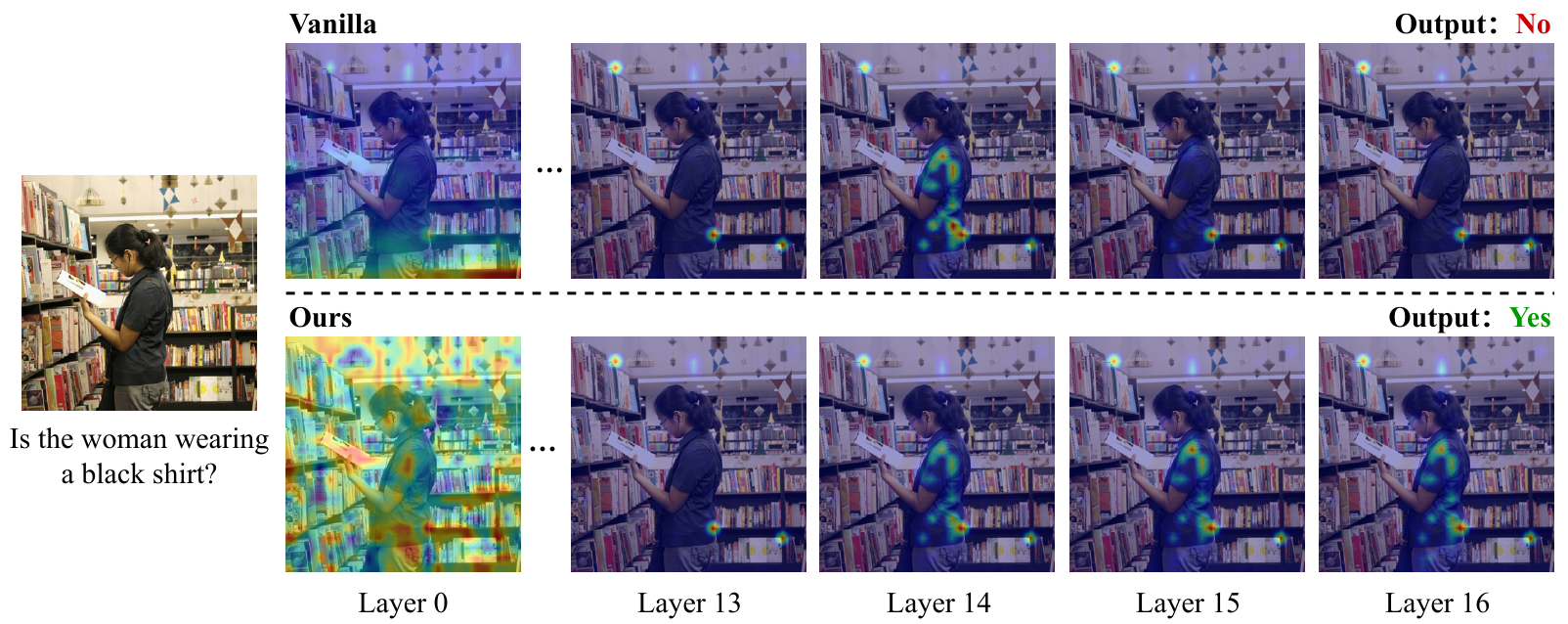}
  \caption{
  A running example of visual attention in LLaVA-7B. In vanilla LVLMs, first-layer attention is biased toward the bottom, and focus on the key object (woman) decays after layer 14.
  }
  \label{fig:vis_attn}
\end{figure*}
\begin{figure}[htbp]
  \centering
  \includegraphics[width=\linewidth]{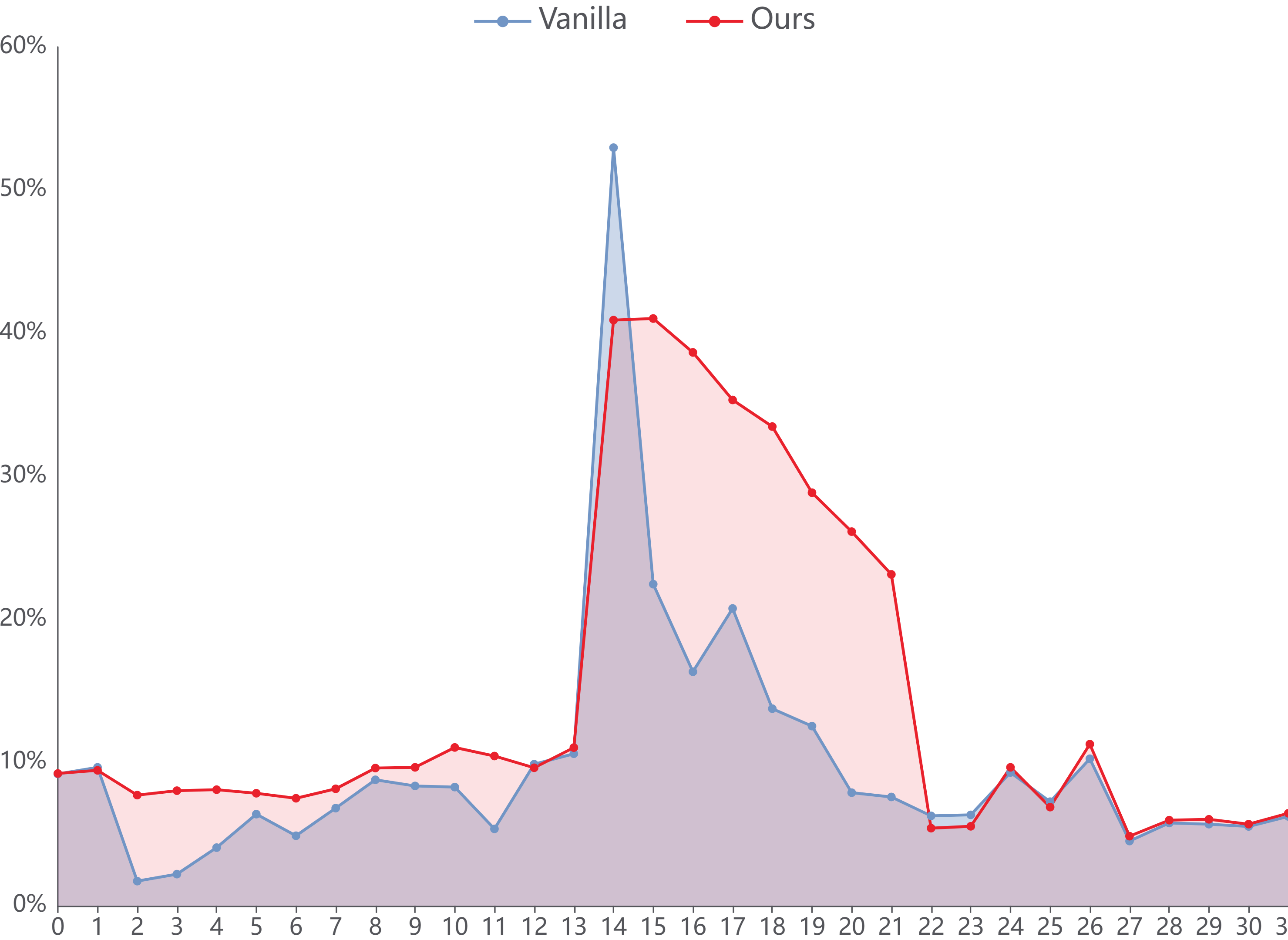}
  \caption{
 Comparison of visual attention to key objects across layers between LVLMs with and without CLVS.
  }
  \label{fig:vis_attn_all}
\end{figure}

While previous studies have demonstrated effectiveness in improving the visual attention of LVLMs, they primarily focus on enhancing attention within individual layers without considering the cross-layer evolution of visual attention.
We observe that visual attention can change dramatically between consecutive layers. For example, in \cref{fig:vis_attn}, when asked ``Is the woman wearing a black shirt?'', the model attends to the ``woman'' and ``shir'' (\emph{i.e.}, key objects) at layer 14 but ceases to focus on them afterward.
Although the model accurately locates key objects, its attention to these objects is very brief. This phenomenon, commonly observed across different LVLMs and images, is referred to as \emph{advantageous attention decay}. We hypothesize that such brief attention is insufficient for effective visual information fusion in LVLMs.
Therefore, sustaining an appropriate focus on key objects is beneficial to enhance the model’s visual capabilities.

To tackle advantageous attention decay, we propose Cross-Layer Vision Smoothing (CLVS) to enhance LVLMs’ visual comprehension by maintaining sustained focus on key objects in the image. To this end, we introduce a vision memory for smoothing visual attention across layers.
Specifically, following previous work~\cite{IMCCD}, we unify the visual attention in the first layer for unbiased perception and initialize the vision memory with the visual attention distribution. In subsequent layers, we progressively smooth visual attention by interpolating it with the vision memory and updating the memory iteratively.
Moreover, since visual understanding primarily occurs in the early and middle layers of LVLMs, we propose an uncertainty-based criterion to determine whether the visual understanding process has been completed and, if so, terminate visual attention smoothing. With CLVS, the model can sustain focus on key objects, thereby improving the visual capabilities of LVLMs.
The comparison of visual attention to key objects across layers between LVLMs with and without CLVS is shown in \cref{fig:vis_attn_all}.

% We conduct extensive experiments on four visual understanding benchmarks and three LVLMs. The results demonstrate that CLVS effectively improves the visual capabilities of LVLMs, reducing hallucinations, particularly those related to object attributes and relations.

In summary, our contributions are:
\begin{itemize}
\item We analyze the cross-layer evolution of visual attention and identify advantageous attention decay, a phenomenon where the model’s focus on key objects is short-lived.
\item We propose Cross-Layer Vision Smoothing (CLVS), which sustains the model’s focus on key objects to enhance visual understanding.
\item We conduct extensive experiments and in-depth analyses to show that CLVS effectively alleviates advantageous attention decay and reduces hallucinations, particularly those related to object attributes and relations.
\end{itemize}

\section{Related Work}
\label{sec:related work}

The cross-modal understanding ability of LVLMs is built upon the attention mechanism. However, end-to-end training lacks supervision over the model’s internal processes, leading to shortcomings in the spontaneously formed visual attention, such as position bias and visual attention sink. These issues can result in hallucinations~\cite{hallu_lvlm, CICD} in the model’s output.

Numerous works have explored improving visual attention in LVLMs. \citet{AD-HH, DAC} incorporate supervisory signals into the attention mechanism, enabling fine-grained control during training.
To avoid high training costs and improve generalizability, an increasing number of studies focus on training-free approaches to optimize visual attention.
Given the critical role of visual information in LVLMs, \citet{PAI, IBD, VAF} directly increase the model’s attention weights on visual tokens. While effective, these approaches overlook the internal dynamics of visual attention.
Leveraging the diversity of multi-head attention, \citet{EAH, SEVI, AD-HH} identify and utilize high-quality attention heads to optimize visual attention.
Recent studies~\cite{VAR, TARAC} further analyze attention distributions to identify the causes of hallucinations and implement targeted optimizations. However, these works enhance visual attention independently within each layer, overlooking its evolution across layers. 

Different from the above approaches, we propose CLVS to maintain a sustained focus on key objects throughout the model. Our method introduces a vision memory to smooth attention distributions across layers, ensuring continuous focus on key objects and enhancing the visual capabilities of LVLMs.

\section{Cross-Layer Vision Smoothing}
\begin{figure}[tbp]
  \centering
  \includegraphics[width=\linewidth]{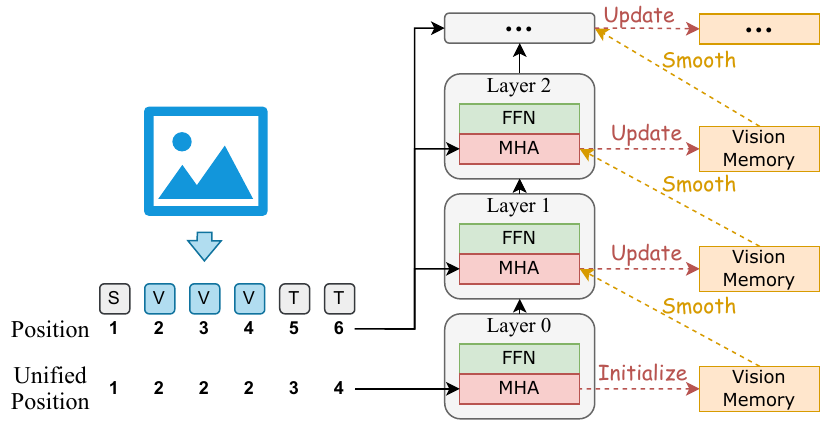}
  \caption{
  Overview of CLVS. \emph{S}, \emph{V}, and \emph{T} denote system tokens, visual tokens, and textual tokens respectively. Unified position indices are applied in the first layer. The vision memory, initialized from this unbiased visual attention, smooths and is updated across subsequent layers.}
  \label{Fg_diagram}
\end{figure}

To ensure sustained attention to key objects in the image, we propose Cross-Layer Vision Smoothing (CLVS). Figure~\ref{Fg_diagram} provides an overview of our method. We first normalize the positional relationships between visual tokens in the initial layer, then initialize a vision memory with unbiased visual attention. This vision memory smooths the visual attention and is dynamically updated across subsequent layers. Finally, the smoothing mechanism is terminated once the model completes the visual understanding process.

\subsection{Unified Visual Positions}
Given a Transformer-based LVLM $f_{\bm{\theta}}$ with parameters $\bm{\theta}$, the goal of the model is to predict the next-token distribution given textual inputs $\mathbf{x}$, visual inputs $\bm{v}$, and previously generated texts $\mathbf{y}{<t}$, which can be formulated as
\begin{align}
p(y_t \mid \mathbf{x}, \bm{v}, \mathbf{y}_{<t}) = f_{\bm{\theta}}(\mathbf{x}, \bm{v}, \mathbf{y}_{<t}).
\end{align}
The LVLM employs an embedding layer and a vision encoder to map both text and images into a sequence of vectors. Positional embeddings are then added to establish contextual relationships among the tokens.

Positional embeddings are determined by the absolute or relative position of a token in a sequence and play an important role in Transformer-based models. Recent studies~\cite{DAC, IMCCD} have found that Transformer-based LVLMs tend to focus disproportionately on the bottom-right corner of the image (\emph{i.e.}, the visual tokens at the end of the input sequence), a bias influenced by the positions of visual tokens in the inputs. Unlike textual inputs, visual information does not inherently possess sequential characteristics. Therefore, the model should receive unbiased attention, particularly in the first layer, where it has no prior visual perception.

Following previous work~\cite{IMCCD}, we normalize the positional indices of all image tokens to a unified index and replace the original positional indices in the concatenated input sequence. Given a sequence of length $N = N_s + N_v + N_i$, the new positional indices are formally defined as
\begin{equation}
\mathbf{p} = [\{i\}^{N_s}_{i=0}, \{N_s +1\}^{N_v}_{i=0}, \{i+1\}^{(N_s+N_i)}_{i=N_i+1} ],
\end{equation}
where $N_s$, $N_v$, and $N_i$ denote the number of tokens in the system prompt, image, and user prompt, respectively. Note that since we only adjust the visual attention of the current generating token, the unified position does not affect the attention among visual tokens. Because of the causal mask in the attention layer, information flows unidirectionally through the input sequence~\cite{SEVI, VisInfoFlow}, which aligns with the positional indices. Therefore, we apply position debiasing to the input image only in the first layer.

\subsection{Visual Attention Smoothing}
The multi-head attention layers aggregate information from the sequence of vectors into a single representation through a weighted sum. Formally, at decoding step $t$, given a query vector $\bm{q} \in \mathbb{R}^d$ where $d$ is the hidden size of $f$, and a set of key vectors $\bm{K} \in \mathbb{R}^{N \times d}$ where $N$ is the sequence length of all tokens, the attention weights $\bm{\alpha}^{(l)}_h \in \mathbb{R}^N$ computed by head $h$ at layer $l$ are defined as
\begin{align}
\bm{\alpha}_h^{(l)} = \mathrm{softmax}(\frac{\bm{q}\bm{K}^\mathsf{T}}{\sqrt{d}}).
\end{align}

Previous studies \cite{ViCrop, AdaptVis} reveal a strong correlation between the visual understanding capabilities of LVLMs and their visual attention. However, we found that the model’s attention quickly converges on sink tokens, hindering sustained focus on key objects. To mitigate this issue and ensure stable attention on key objects, we introduce a vision memory $\bm{m}$ into the LVLM, which serves as a cross-layer smoothing mechanism in visual attention. Let $\bm{\alpha}^{(l)}_h = \{\bm{\lambda}^{(l)}_h, \bm{\mu}^{(l)}_h\}$, where $\bm{\lambda}^{(l)}_h \in \mathbb{R}^{N_v}$ denotes the attention weights associated with visual tokens ($N_v$ is the number of visual tokens), and $\bm{\mu}^{(l)}_h \in \mathbb{R}^{N - N_v}$ denotes the attention weights associated with non-visual tokens. The vision memory $\bm{m}$ stores only the attention weights corresponding to visual tokens. We initialize the vision memory at the first layer, denoted as $\bm{m}^{(1)}$, using the unbiased visual attention of that layer, and apply max pooling across attention heads to compress the model’s overall attention to the image:
\begin{equation}
\bm{m}^{(1)}[i] = \max_{1 \le h \le H} \bm{\lambda}^{(1)}_h[i],
\end{equation}
where $\bm{m}^{(1)}[i] \in \mathbb{R}$ denotes the $i$-th element of $\bm{m}^{(1)}$, $H$ is the total number of attention heads, and $i \in \{1, \ldots, N_v\}$. This highlights the most salient visual information captured at each layer and propagates it across different attention heads.

In the subsequent Transformer layers, we smooth the visual attention weights for each head and update the vision memory iteratively. The smoothing formula for visual attention is formally defined as
\begin{equation}\label{Eq_beta}
    \bm{\hat{\lambda}}_h^{(l)} =  \beta \bm{\lambda}^{(l)}_h + (1-\beta) \bm{m}^{(l-1)},\quad 1\le h \le H,
\end{equation}
where $\beta \in [0,1]$ is a hyperparameter that controls the extent of smoothing. However, smoothing the visual attention weights may invalidate the normalization property of $\bm{\alpha}^{(l)}_h$, so we need to re-normalize the attention scores. Since directly re-normalizing the attention distribution with the $\mathrm{softmax}$ function may reduce its sharpness, we instead apply a simple re-normalization by:
\begin{equation}
    \bm{\hat{\alpha}}^{(l)}_h [i] = \frac{\bm{\alpha}^{(l)}_h[i]}{\sum_{k=1}^n \bm{\alpha}^{(l)}_h[k]},\quad 1\le h \le H.
\end{equation}
The vision memory is then updated layer by layer based on the original visual attention from each layer:
\begin{equation}\label{Eq_gamma}
    \bm{m}^{(l)}[i] =  \gamma \bm{m}^{(l-1)}[i] + (1-\gamma) \max_{1 \le h \le H} \bm{\lambda}^{(l)}_h[i],
\end{equation}
where $\gamma \in [0, 1]$ is the soft memory window size that limits how far the model can look back into previous vision memories. A larger $\gamma$ allows the model to access more distant vision memories across layers.

Since the vision memory is initialized with unbiased visual attention, our method allows the model to capture more visual information in the early layers. Once the model identifies key objects, the smoothing mechanism slows the decay of visual attention, enabling sustained focus on those objects.

\subsection{Vision Understanding Completion}
The visual understanding process in LVLMs primarily occurs in the early and middle layers, while language-level feature construction takes place in the final layers~\cite{VisInfoFlow, SEVI}. Therefore, adjustments to visual attention should target the early and middle layers. Previous works~\cite{TARAC, VAF} often determine the layer range for attention optimization by setting fixed hyperparameters, which limits generalization across different LVLMs. To address this, we propose a dynamic mechanism that determines the active layer range by assessing whether the model’s visual understanding process is complete, and terminates CLVS accordingly.

% When the model's output token remains unchanged across all subsequent layers, i.e., the output has reached a converged state, we consider the visual understanding process to be complete.
Intuitively, if the model’s output has converged (\emph{i.e.}, the output token remains unchanged in subsequent layers under greedy decoding), we can terminate the smoothing process. However, output convergence is determined based on the final layer’s output and therefore cannot be known at the current layer.
A straightforward approach would be to perform a preliminary forward pass to determine the convergence status at each layer, but this significantly increases inference cost.
We instead use output uncertainty~\cite{LookTwice} at each layer as a surrogate measure to determine the end of visual understanding. 
Specifically, we multiply the representation by the unembedding matrix to obtain the probability distribution at the current layer $l$, and calculate the uncertainty measure $u$ as
\begin{equation}
  u = \frac{\sum_{k=1}^{|V|} -p_t \log p_t}{\log |V|},
\end{equation}
where $|V|$ denotes the vocabulary size.
For computational concerns, we compute only the top-$10$ tokens (\emph{i.e.}, $|V|=10$).

\begin{figure}
  \centering
  \begin{subfigure}{0.49\linewidth}
    % \fbox{\rule{0pt}{2in} \rule{.9\linewidth}{0pt}}
    \includegraphics[width=\linewidth]{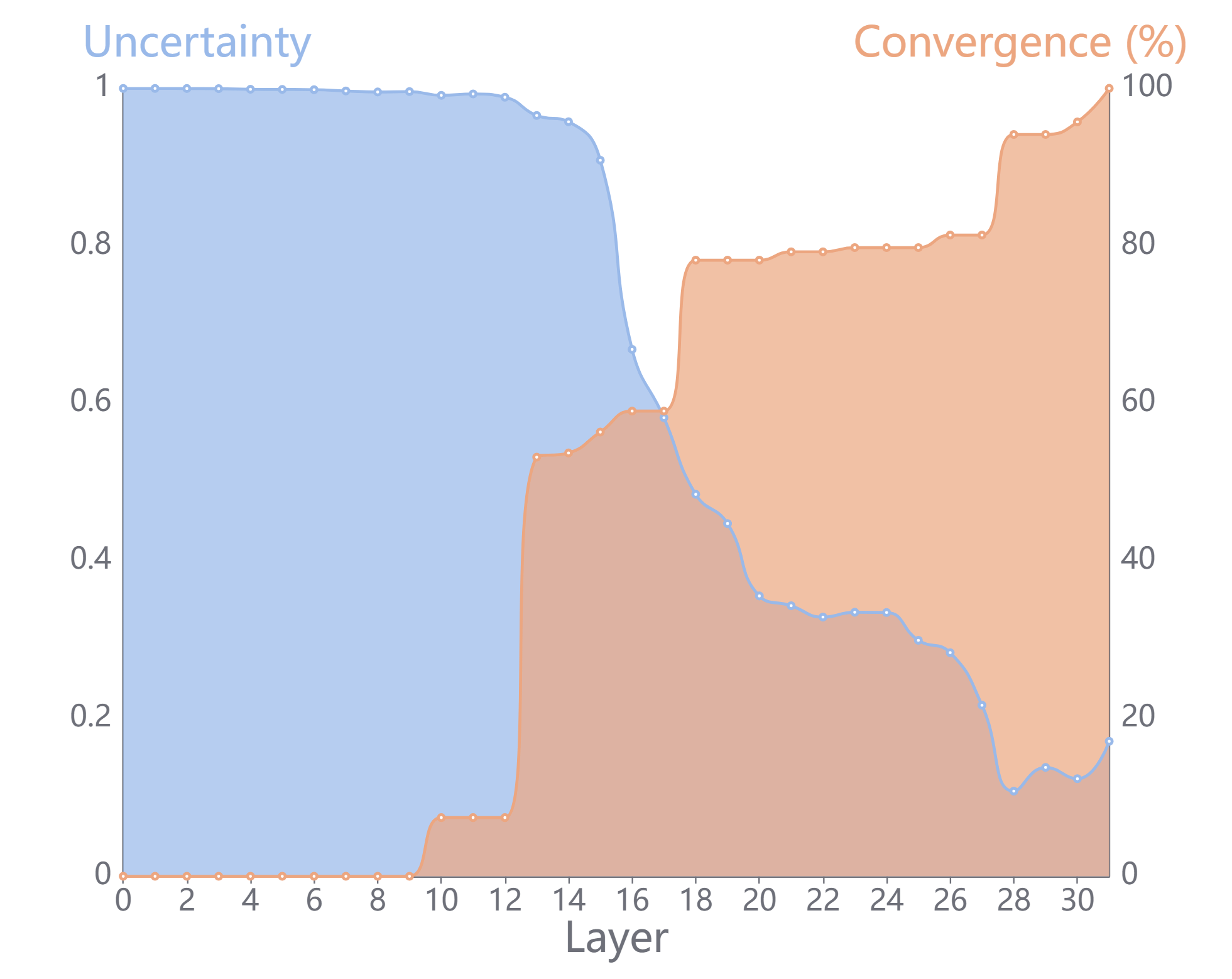}
    % \caption{.}
    % \label{fig:vsc_pfm-a}
  \end{subfigure}
  \hfill
  \begin{subfigure}{0.49\linewidth}
    % \fbox{\rule{0pt}{2in} \rule{.9\linewidth}{0pt}}
    \includegraphics[width=\linewidth]{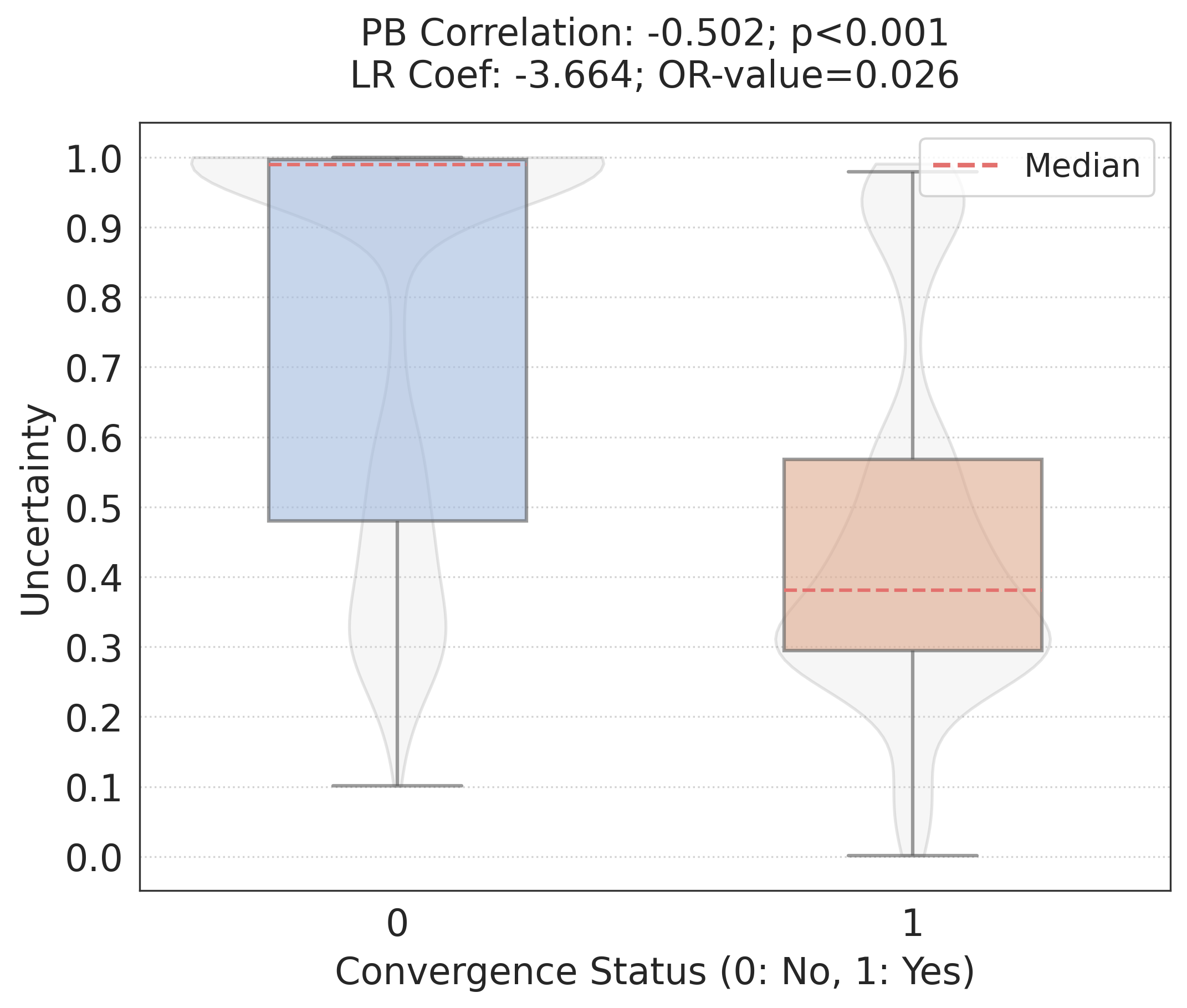}
    % \caption{Another example of a subfigure.}
    % \label{fig:vsc_pfm-b}
  \end{subfigure}
    \caption{\label{fig:unc}
  Statistical results of output uncertainty and convergence on the MME benchmark. Left: clear inverse relationship between convergence and uncertainty. Right: \emph{PB Correlation} denotes Point-Biserial Correlation, and \emph{LR Coefficient} denotes the coefficient from Logistic Regression analysis.
  }
\end{figure}

% \begin{figure}[t]
%   \centering
%   \subfloat{
%   \begin{minipage}[t]{0.5\linewidth}
%   \centering
%   \includegraphics[width=\textwidth]{figures/unc_conv.png}
%   %\caption{fig1}
%   \end{minipage}%
%   }%
%   \subfloat{
%   \begin{minipage}[t]{0.5\linewidth}
%   \centering
%   \includegraphics[width=\textwidth]{figures/cor.png}
%   %\caption{fig2}
%   \end{minipage}%
%   }%
%   \centering
%   \caption{\label{fig:unc}
%   Statistical results of output uncertainty and convergence on the MME benchmark. Left: clear inverse relationship between convergence and uncertainty. Right: \emph{PB Correlation} denotes Point-Biserial Correlation, and \emph{LR Coefficient} denotes the coefficient from Logistic Regression analysis.
%   }
% \end{figure}

We analyze the uncertainty and convergence of the model’s responses, with results shown in \cref{fig:unc}.
From the left panel, we observe a strong correlation between convergence and uncertainty: as uncertainty decreases, the model progressively completes visual perception and begins constructing semantic information.
Both point-biserial correlation and logistic regression analyses, shown in the right panel, further confirm statistical associations between these measures. Based on these observations, we use uncertainty to determine whether the model has completed visual perception. 
Once the uncertainty falls below the threshold $\delta$, Visual attention smoothing is terminated in subsequent Transformer layers.

\begin{table*}
 \setlength{\tabcolsep}{3mm}
\centering
\caption{\label{tab:AMBER}
Results on AMBER. \textbf{Bold} indicates the best result in each setting, and \underline{underline} indicates the second best. ``All'' denotes the aggregated results over all test samples.
}
\begin{tabular}{l|l|cc|cc|cc|cc}
\toprule
\multirow{2}{*}{\textbf{Model}} & \multirow{2}{*}{\textbf{Method}} 
& \multicolumn{2}{c|}{\textbf{Existence}} 
& \multicolumn{2}{c|}{\textbf{Attribute}} 
& \multicolumn{2}{c|}{\textbf{Relation}} 
& \multicolumn{2}{c}{\textbf{All}} \\
\cmidrule(lr){3-4} \cmidrule(lr){5-6} \cmidrule(lr){7-8} \cmidrule(lr){9-10}
 & & Acc.  $\uparrow$ & F1 $\uparrow$ & Acc. $\uparrow$ & F1 $\uparrow$ & Acc. $\uparrow$ & F1 $\uparrow$ & Acc. $\uparrow$ & F1 $\uparrow$ \\
\midrule

\multirow{8}{*}{LLaVA-7B} 
  & Vanilla & 71.4  & 83.3 & 72.0  & 64.4 & 74.1  & 68.5 & 72.1  & 74.8 \\
  & VCD &  68.3 & 81.1 & \underline{72.7}  & \underline{66.6} & 72.4  & 65.1 & 71.1  & 74.1 \\
  & PAI  & 70.3 & 82.5 & 69.5 & 59.1 & 74.5 & 65.6 & 70.3 & 72.5\\
  & PAI$_{\mathrm{CD}}$ & \underline{75.7} & \underline{86.1} & 71.9  & 64.2 & \underline{75.8}  & \underline{71.2} & \underline{73.7} & \underline{76.7} \\
  & VAR& 70.7 & 82.8 & 72.1  & 65.0 & 73.7  & 66.6 &  71.8  & 74.6 \\
  & AD-HH & 71.4 & 83.3 & 71.8  & 64.2 & 74.0  & 68.7 & 71.9  & 74.8\\
  & TARAC & 72.1  & 83.7 & 71.4  & 63.5 & 74.2  & 68.4 & 72.0  & 74.8 \\
  \rowcolor{gray!20}
  & CLVS &  \textbf{80.1} & \textbf{88.9} & \textbf{73.7} & \textbf{67.3} & \textbf{76.7} & \textbf{74.9} & \textbf{76.2} & \textbf{79.6} \\

\midrule

\multirow{8}{*}{LLaVA-13B} 
  & Vanilla & 66.0 & 79.5 & 75.3  & 69.5 & 70.1  & 45.2 & 71.5 & 73.5 \\
  & VCD    & 64.6 & 78.4 & 75.9  & 71.3 & 69.9  & 45.7 & 71.3  & 73.5 \\
  & PAI  & 69.2 & 81.7 & 77.5 & 73.2 & \underline{73.9} & \underline{56.7} & 74.2 & 76.7\\
  & PAI$_{\mathrm{CD}}$  & \underline{70.5}  & \underline{82.6} & \underline{78.1} & \underline{74.3} & \underline{73.9}  & 56.6 & \underline{75.0} & \underline{77.6} \\
  & VAR   & 65.2  & 78.9 & 75.1  & 69.1 & 69.7  & 43.8 & 71.0  & 72.9 \\
  & AD-HH & 67.1  & 80.3 & 75.8  & 70.4 & 70.0  & 45.5 & 72.1  & 74.2 \\
  & TARAC &  68.7  & 81.4 & 76.2  & 70.9 & 71.3  & 48.7 & 73.0  & 75.3\\
  \rowcolor{gray!20}
  & CLVS & \textbf{76.3} & \textbf{86.5} & \textbf{78.7} & \textbf{75.1} & \textbf{74.9} & \textbf{59.9} & \textbf{77.4} & \textbf{80.3} \\

\midrule

\multirow{5}{*}{LLaVA-Next} 
  & Vanilla & \textbf{95.1} & \textbf{97.4} & 84.0  & 85.0 & 67.2  & 71.5 & 85.9 & 89.8\\
  & VCD~\cite{VCD}  & \underline{95.0}  & \textbf{97.4} & 84.4  & 85.2 & 67.7  & 71.8 & 86.1  & \underline{89.9}\\
  & PAI  & 94.9 & \underline{97.3} & \underline{84.6} & \underline{85.3} & \underline{70.0} & \underline{73.2} & \underline{86.4} & \textbf{90.1}\\
  & TARAC~\cite{TARAC}& \underline{95.0}  & \textbf{97.4} & 84.2  & 85.1 & 66.2 & 70.9 & 85.8  & 89.8 \\
  \rowcolor{gray!20}
  & CLVS  &  94.3 & 97.0 & \textbf{85.1} & \textbf{85.6} & \textbf{70.3} & \textbf{73.3} & \textbf{86.6} & \textbf{90.1} \\
  
\bottomrule
\end{tabular}
\end{table*}

\begin{table*}
 \setlength{\tabcolsep}{3mm}
    \centering
    \caption{
    Results on POPE and R-Bench. POPE's results are the average performance across the MSCOCO, A-OKVQA, and GQA datasets.
    }\label{tab:pope}
    \begin{tabular}{l|l|cc|cc|cc|cc}
       \toprule
       \multirow{2}{*}{\textbf{Model}} & \multirow{2}{*}{\textbf{Method}}
       &  \multicolumn{2}{c|}{\textbf{Random}} &  \multicolumn{2}{c|}{\textbf{Popular}} &  \multicolumn{2}{c|}{\textbf{Adversarial}}&  \multicolumn{2}{c}{\textbf{R-Bench}}\\
        \cmidrule(lr){3-4}\cmidrule(lr){5-6}\cmidrule(lr){7-8}\cmidrule(lr){9-10}
       &  &  Acc. $\uparrow$ &  F1 $\uparrow$&Acc.  $\uparrow$&  F1 $\uparrow$&  Acc. $\uparrow$&   F1 $\uparrow$ &  Acc.$\uparrow$ &   F1 $\uparrow$\\
       \midrule
       \multirow{7}{*}{LLaVA-7B} 
       & Vanilla &   87.12  & 87.94  & 79.38  & 82.14  & 71.85  & 77.04  & 67.76 & 74.84    \\
       & VCD &    84.89  & 86.07  & 77.38  & 80.64  & 71.03  & 76.39   & 68.22 & 74.97   \\
       & PAI &    86.20  & 87.20  & 78.45  & 81.56  & 70.83  & 76.46  & 67.17 & 74.66 \\
       & PAI$_\mathrm{CD}$ &   \underline{87.27}  & \underline{87.98}  & \underline{80.03}  & \underline{82.51}  & \underline{72.61}  & \underline{77.38}  & \underline{68.69} & \underline{75.39}   \\
       & AD-HH &    87.13  & 87.93  & 79.43  & 82.18  & 71.89  & 77.07  & 67.60 & 74.76    \\
       & VAF &   84.88  & 86.34  & 77.13  & 80.82  & 69.59  & 75.94   & 66.79 & 74.45    \\
       & TARAC &  86.99  & 87.80  & 79.37  & 82.10  & 71.92  & 77.03  & 67.67 & 74.83     \\
  \rowcolor{gray!20}
       & CLVS &   \textbf{88.74}  & \textbf{88.98}  & \textbf{82.17}  & \textbf{83.76}  & \textbf{75.34}  & \textbf{78.80}   & \textbf{69.01} & \textbf{75.47}     \\
       \midrule
       \multirow{6}{*}{LLaVA-13B} 
       & Vanilla &   84.31  & 85.96  & 78.64  & 81.87  & 71.58  & 77.23 & 68.18 & 75.39     \\
       & VCD &    83.01  & 84.81  & 77.37  & 80.81  & 71.38  & 76.84   &   68.88 & 75.48  \\
       & PAI &    85.36  & 86.71  & 79.72  & 82.56  & 72.38  & 77.64  & 70.31 & 76.51    \\
       & PAI$_\mathrm{CD}$ &  86.10  & 87.24  & \underline{80.86}  & \underline{83.31}  & \underline{73.44}  & \underline{78.26} & \textbf{71.00} & \textbf{76.89}   \\
       & AD-HH &   84.88  & 86.37  & 79.28  & 82.29  & 72.26  & 77.62   &   68.61 & 75.61   \\
       & VAF &    82.42  & 84.62  & 77.10  & 80.91  & 70.08  & 76.41  & 67.66 & 75.10    \\
       & TARAC &   \underline{87.02}  & \underline{87.80}  & 79.40  & 82.10  & 71.95  & 77.03   & 68.79 & 75.71     \\
  \rowcolor{gray!20}
       & CLVS &   \textbf{87.26}  & \textbf{88.08}  & \textbf{82.73}  & \textbf{84.55}  & \textbf{75.09}  & \textbf{79.16}  & \underline{70.80} & \underline{76.79}   \\
       \midrule
       \multirow{4}{*}{LLaVA-Next} 
       & Vanilla &   87.36  & 85.90  & 85.45  & 84.14  & 82.31  & 81.40  & 81.20 & 81.88     \\
       & PAI &   \underline{87.71}  & \underline{86.34}  & \underline{85.95}  & \underline{84.70}  & \underline{82.64}  & \underline{81.82}   &  \textbf{81.88} & \underline{82.57}   \\
       % & VAF &    89.31  & 88.45  & 86.58  & 85.95  & 83.14  & 83.00   \\
       & TARAC &   86.13  & 84.26  & 83.94  & 82.25  & 81.04  & 79.73 & \underline{81.76} & 82.33   \\
  \rowcolor{gray!20}
       & CLVS &   \textbf{88.69}  & \textbf{87.59}  & \textbf{86.21}  & \textbf{85.31}  & \textbf{82.80}  & \textbf{82.36}  & 81.67 & \textbf{82.60}     \\
       \bottomrule
    \end{tabular}
\end{table*}

\begin{table*}[t]
    \setlength{\tabcolsep}{3mm}
    \centering
    \caption{Results on image captioning task.}
    \label{tab:cap}
    \begin{tabular}{l|l|cccc|ccc}
       \toprule
       \multirow{2}{*}{\textbf{Model}} & \multirow{2}{*}{\textbf{Method}} &
       \multicolumn{4}{c|}{\textbf{AMBER}} & \multicolumn{3}{c}{\textbf{CHAIR}} \\
       \cmidrule(lr){3-6}\cmidrule(lr){7-9}
       &  & CHAIR $\downarrow$& Cover $\uparrow$& Hal  $\downarrow$& Cog  $\downarrow$& CHAIRs  $\downarrow$& CHAIRi  $\downarrow$& F1 $\uparrow$\\
       \midrule
       \multirow{5}{*}{LLaVA-7B} 
       & Vanilla        &  7.1 & \textbf{50.7} & 32.5 & 3.8 & 52.2 & 13.9 & 76.1 \\
       & PAI            &  5.8 & 48.8 & 27.5 & \underline{2.5} & 33.6 & \underline{9.0} & \textbf{77.1} \\
       & PAI$_{\mathrm{CD}}$ &  \underline{5.1} & 48.4 & \underline{25.7} & \textbf{2.1} & \textbf{30.8} & \textbf{8.1} & \underline{76.3} \\
       & TARAC          &  5.8 & \underline{49.3} & 28.5 & 2.9 & 43.0 & 11.2 & 75.1 \\
  \rowcolor{gray!20}
       & CLVS           &  \textbf{4.2} & 48.0 & \textbf{21.9} & \textbf{2.1} & \underline{33.0} & 9.2 & 75.9 \\
       \midrule
       \multirow{5}{*}{LLaVA-13B} 
       & Vanilla        &  6.8 & 51.3 & 31.1 & 3.4 & 52.6 & 14.1 & 76.4 \\
       & PAI            &  5.4 & \underline{51.6} & \underline{27.3} & 2.1 & \underline{40.2} & \underline{10.8} & 76.9 \\
       & PAI$_{\mathrm{CD}}$ &  \underline{5.2} & \textbf{52.0} & 28.3 & \textbf{1.9} & \textbf{37.0} & \textbf{9.9} & \textbf{77.3} \\
       & TARAC          &  5.8 & 49.3 & 28.5 & 2.9 & 43.0 & 11.2 & 75.1 \\
  \rowcolor{gray!20}
       & CLVS           &  \textbf{4.9} & 50.9 & \textbf{23.7} & \underline{2.0} & 43.8 & 11.5 & \underline{77.0} \\
       \midrule
       \multirow{4}{*}{LLaVA-Next} 
       & Vanilla        &  7.8 & \underline{63.1} & 46.1 & 3.9 & 34.2 & 9.3 & \underline{72.2} \\
       & PAI            &  \underline{7.1} & \textbf{64.2} & \underline{42.2} & \textbf{3.2} & 34.0 & 8.9 & \textbf{73.7} \\
       & TARAC          &  \textbf{6.7} & 60.3 & \textbf{37.5} & \underline{3.4} & \textbf{29.8} & \textbf{8.0} & 71.5 \\
  \rowcolor{gray!20}
       & CLVS           &  \underline{7.1} & 62.4 & 42.9 & \underline{3.4} & \underline{32.2} & \underline{8.2} & 71.7 \\
       \bottomrule
    \end{tabular}
\end{table*}

\section{Experiments}
\subsection{Setup}

\paragraph{Benchmarks.}
We use the following benchmarks to test the effectiveness of our method:
\begin{itemize}
\item \textbf{AMBER}~\cite{AMBER}: a human-annotated benchmark that includes both visual question answering (VQA) and image captioning tasks. The VQA task encompasses three types of hallucination: existence, attribute, and relation. The image captioning task evaluates object hallucination.
\item \textbf{R-Bench}~\cite{R-Bench}: a benchmark for evaluating relation hallucination, which pose a greater challenge to the model's visual understanding ability.
This benchmark provides multiple randomly constructed subsets with a balanced number of \emph{Yes} and \emph{No} questions, and we use the subset with ID 1.
We conduct experiments using image-level questions.
\item \textbf{POPE}~\cite{POPE}: a benchmark for evaluating existence hallucination. It consists of three subsets: \emph{Random}, \emph{Popular}, and \emph{Adversarial}, each constructing negative examples in a distinct manner.
\item \textbf{CHAIR}~\cite{chair}: a benchmark for evaluating object hallucination in image captioning, assessing captions by annotations from the MSCOCO dataset.
\end{itemize}
% We also conduct experiments on the following benchmarks to further validate our proposed CLVS method:
% \begin{itemize}

% \item \textbf{MME}~\cite{MME}: a comprehensive evaluation benchmark. We use perception tasks to test the model's visual understanding ability following previous works, which includes five types of questions: \emph{Existence}, \emph{Color}, \emph{Count}, \emph{Position}, and \emph{OCR}.
% \end{itemize}

\paragraph{Baselines.}
We compare our method with the following baselines:
\begin{itemize}
\item \textbf{VCD}~\cite{VCD}: a method that stimulates visual uncertainty by adding noise to images and purifies visual information using contrastive decoding.
\item \textbf{PAI}~\cite{PAI}: a method that directly amplifies visual attention and performs contrastive decoding with pure textual context to purify visual information. To evaluate the effect of enhanced visual attention, we denote the variant without contrastive decoding as PAI and the variant with contrastive decoding as  PAI$_\mathrm{CD}$.
% To test We denote the variants with and without contrastive decoding as PAI$_\mathrm{CD}$ and PAI, respectively. 
\item \textbf{VAR}~\cite{VAR}: a method that identifies visual sink tokens and redistributes the model's attention from sink tokens to other visual tokens.
\item \textbf{AD-HH}~\cite{AD-HH}: a method that identifies attention heads that cause hallucinations. When the attention of these heads to text exceeds a threshold, their attention to text is set to zero, thereby enhancing visual attention.
% \item \textbf{VAF}~\cite{VAF}: a method that redistributes the model's attention from system tokens to visual tokens in the middle layers where the model performs visual understanding.
\item \textbf{TARAC}~\cite{TARAC}: a method that optimizes visual attention in the middle layers. It accumulates visual attention from each generation step and injects it into the visual attention of the current step.
\end{itemize}
% The implementation details of baselines are provided in the Appendix.

\paragraph{Implementation Details.}
In all experiments, LVLMs use greedy search for next-token prediction.
We set $\beta = 0.8$ in \cref{Eq_beta} and $\gamma=0.8$ in \cref{Eq_gamma} for all LVLMs.
The uncertainty threshold $\delta$ is set to 0.5 for LLaVA-7B and LLaVA-13B, 0.9 for LLaVA-Next.
We conduct our experiments on a single NVIDIA A800 40G GPU. 
% More details are given in the supplementary material.

\subsection{Results on Visual Understanding Task}
\paragraph{Results on AMBER.} \Cref{tab:AMBER} shows the results on AMBER, a comprehensive hallucination evaluation benchmark that includes tests for three major types of hallucinations.
% Our method achieves notable improvements over baselines on multiple LVLMs, particularly in understanding object relationships.
% Our method achieves consistent improvements across all three tasks, with the most significant gains observed in relation hallucination.
A series of baselines based on visual attention optimization achieve significant improvements over the Vanilla results, highlighting the importance of visual attention in enabling LVLMs to understand visual content. PAI$_\mathrm{CD}$ achieves the second-best performance across all experiments; however, it requires two forward passes for contrastive decoding, doubling the inference cost. In contrast, our method achieves better performance with only a single forward pass, introducing no significant increase in computational overhead.
Additionally, we observe that CLVS does not achieve the best performance on the existence hallucination task in the LLaVA-Next experiments. This is because the base model's performance is already close to the upper bound of the dataset, and we notice some annotation noise in the data—making it difficult to determine ground truth accurately—which may lead to performance fluctuations. To further evaluate CLVS on existence hallucination, we additionally select the POPE benchmark for evaluation.
% Compared with other tasks, the improvement in determining object existence is relatively small, especially when using models such as Qwen2.5-VL and LLaVA-Next, which possess stronger visual comprehension capabilities. This is because our approach leverages the inherent attention advantages of LVLMs to maintain focus on key objects, thereby substantially enhancing performance in complex relational reasoning. However, in object-existence determination tasks where relevant objects may be absent from the image, our method cannot establish focused attention on key objects and thus provides limited improvements.
%We will provide a more thorough discussion of this phenomenon on the relation hallucination benchmark R-Bench and the existence hallucination benchmark POPE.
\paragraph{Results on POPE.} 
We further evaluate CLVS on existence hallucination, and the results are shown in the left part of \cref{tab:pope}.
Our method achieves consistent and significant improvements across all three base models, and outperforms all other strategies under various negative sample selection schemes, demonstrating its effectiveness in enhancing the model’s perceptual awareness of objects.
Additionally, we observe that CLVS significantly outperforms other visual attention optimization methods on existence hallucination, which strongly validates the importance of sustained visual focus for robust visual understanding.
% As discussed earlier, our method provides limited improvements when key objects are absent from the image. For example, when the answer to the question ``\emph{Is there a/an \{object\} in the image?}'' is ``\emph{False}'', there are no relevant key objects to attend to, which limits the effectiveness of our approach. To further examine this issue, we conduct additional experiments on the POPE dataset, a benchmark designed for existence hallucination. The results are presented in \cref{tab:pope}. We find that while our method does not yield substantial gains in accuracy, it achieves a notable increase in recall. This indicates that when key objects are present, our method effectively enhances the model’s visual understanding by sustaining attention on them.

\paragraph{Results on R-Bench.}
In the AMBER benchmark, CLVS shows strong performance in understanding object relationships. To further evaluate its effectiveness in enhancing model comprehension of object relations, we conduct experiments on the R-Bench benchmark, with results presented in the right part of \cref{tab:pope}.
CLVS achieves the overall best performance in this Benchmark. 
Although it is slightly lower than  PAI$_\mathrm{CD}$ on LLaVA-13B, it still surpasses PAI, demonstrating that our method retains a clear advantage in enhancing the model’s visual understanding capabilities.

% surpassing PAI and remaining competitive with PAI$_\mathrm{CD}$—even though it is slightly lower than PAI$_\mathrm{CD}$ on LLaVA-13B. This indicates that our method still holds a clear advantage in enhancing the model’s visual understanding capabilities.
% \Cref{tab:pope} shows the results on the R-Bench benchmark. Our method achieves significant improvements in evaluating relational hallucinations, attaining state-of-the-art F1 scores across all LVLMs, including strong models such as Qwen2.5-VL and LLaVA-Next. These results are consistent with those on AMBER, further demonstrating the versatility of our method.

\begin{figure}[t]
  \centering
  \includegraphics[width=\linewidth]{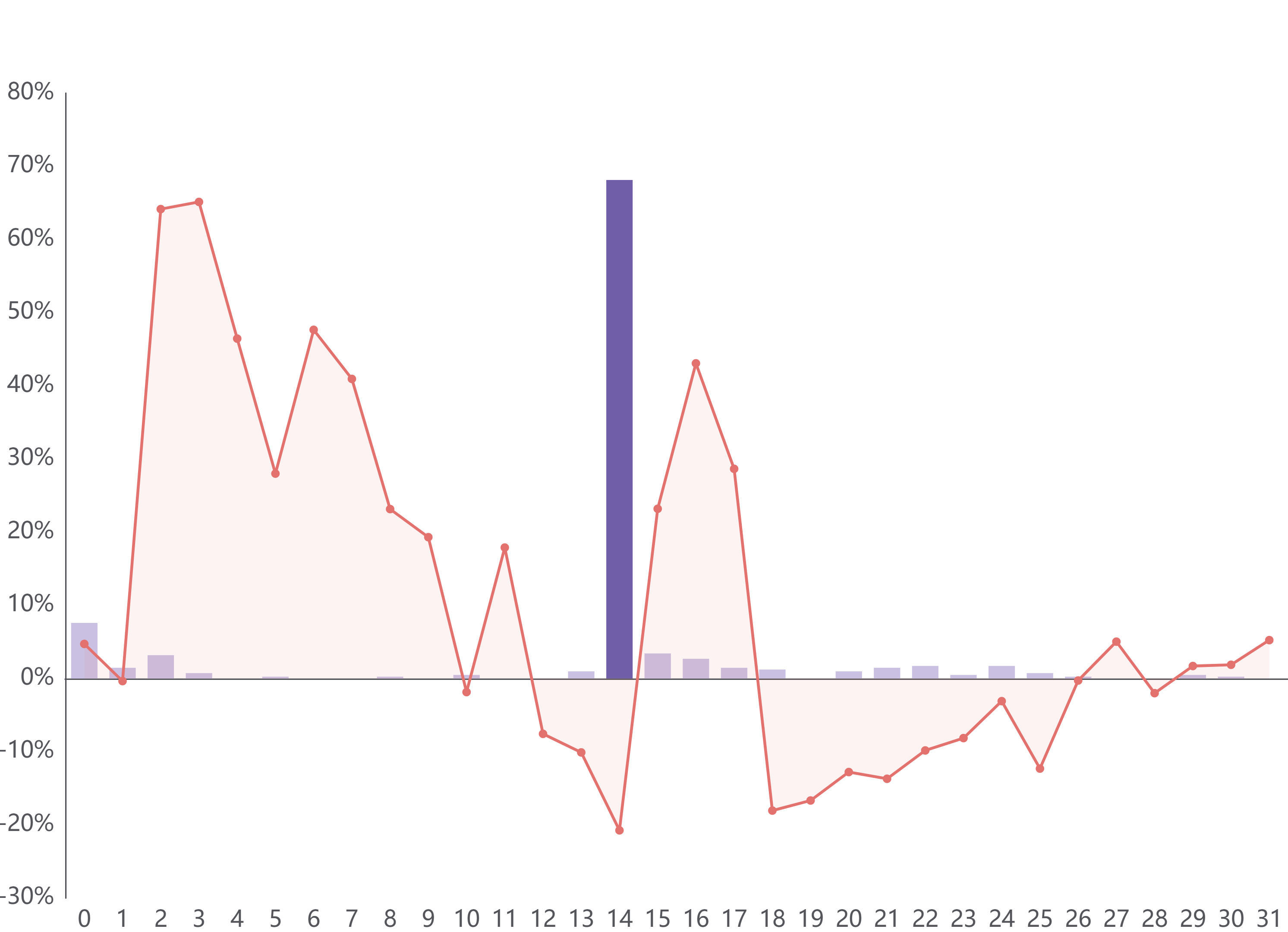}
  \caption{\label{fig:coco}
  Statistical results of visual attention on key objects in LLaVA-1.5. The bar chart shows the distribution of attention peaks across layers in the vanilla method, while the line chart shows the average increase in attention on key objects by CLVS relative to vanilla in each layer.
  }
\end{figure}

\subsection{Results on Image Captioning Task}
CLVS enhances visual understanding by extending the model’s visual attention, and its effectiveness is thoroughly validated across multiple visual understanding benchmarks.
Unlike visual understanding tasks, image captioning requires not only accurate visual perception but also coherent integration of textual features.
To evaluate CLVS’s performance on image captioning, we conduct experiments on two dedicated captioning benchmarks. The results are presented in \cref{tab:cap}.

CLVS significantly reduces hallucinations in image captioning, achieving substantial improvements over the Vanilla baseline on both captioning benchmarks and matching the performance of state-of-the-art methods.

The reason CLVS achieves suboptimal performance in image captioning is that, during the process of generating descriptions, not every token requires visual information. When generating subwords, punctuation, or function words—tokens with strong language priors—the model does not rely on visual attention. In such cases, extending visual attention provides limited benefit.
Despite this, sustained attention remains beneficial when the model generates content words, leading to the significant improvements achieved by CLVS.

% \subsection{Results on MME}
% \begin{figure}[t]
%   \centering
%   \includegraphics[width=0.8\linewidth]{figures/MME.png}
%   \caption{\label{fig:mme}
%   Radar chart of results on the MME benchmark.
%   }
% \end{figure}
% We further evaluate our method on a broader range of tasks using the MME benchmark, with results shown in \cref{fig:mme}. Our method achieves the highest overall score among all baselines except on the position task. We conjecture that the position task relies heavily on spatial information in the image, while our approach primarily prolongs LVLMs’ focus on key objects and therefore does not enhance their understanding of spatial relationships. Nevertheless, our method yields significant improvements on the color and count tasks, where key objects are present.

\begin{table*}
    \centering
    \caption{
    Ablation study on AMBER with LLaVA-7B. ``\emph{w/o} Early Termination'' represents performing vision smoothing until the final layer. ``\emph{w/o} Unified Position'' represents employing the original position indices in the first layer.
     ``TTFT'' represents ``Time to First Token''.
     The AMBER metric is calculated as $(1-\mathrm{CHAIR}+\mathrm{F1})/2$.
    }\label{tab:ablation}
    \begin{tabular}{l|cccc|cccc|c|c}
       \toprule
       % Method & CHAIR $\downarrow$ & Cover $\uparrow$ & Hal $\downarrow$ & Cog $\downarrow$ & Acc. $\uparrow$ & Prec. $\uparrow$ & Rec. $\uparrow$ & F1 $\uparrow$ & AMBER $\uparrow$ & TTFT $\downarrow$ \\
       Setting & CHAIR  & Cover  & Hal  & Cog & Acc.  & Prec.  & Rec.  & F1  & AMBER & TTFT \\
       \midrule
       Vanilla & 7.1 & \textbf{50.7} & 32.5 & 3.8 & 72.1 & 92.2 & 63.0 & 74.8 & 34.35 & \textbf{$\times1.000$} \\
       CLVS & \textbf{4.2} & 48.0 & \textbf{21.9} & 2.1 & 76.2 & 92.5 & 69.9 & 79.6 & \textbf{38.20} & $\times1.035$ \\
       \emph{w/o} Early Termination & 5.4 & 47.7 & 24.1 & 1.9 & \textbf{76.3} & 92.4 & \textbf{70.0} & \textbf{79.7} & 37.65 & $\times1.111$ \\
       \emph{w/o} Unified Position & \textbf{4.2} & 48.3 & 22.9 & \textbf{1.7} & 74.6 & \textbf{93.4} & 66.4 & 77.6 & 37.20 & $\times1.031$ \\
       \bottomrule
    \end{tabular}
\end{table*}

\subsection{Key Object Attention Analysis.} 
To validate the effectiveness of CLVS in strengthening advantageous visual attention, we conduct experiments on 500 positive samples from POPE and use manually annotated instance segmentations from the MSCOCO dataset to localize key objects. We measure the attention to a key object in layer $l$ as follows:
\begin{equation}
    a_{\textrm{obj}}^{(l)} = \sum_{i=1}^{N_v} \bm{a}^{(l)}[i] \cdot \frac{\mathrm{Area}(\textrm{patch}_i \cap \textrm{obj})}{\mathrm{Area}(\textrm{patch}_i)}
\end{equation}
where $N_v$ is the number of visual tokens and $\bm{a}^{(l)}[i]$ is the average attention value for visual token $i$ across attention heads. We analyze the peak-layer distribution $l_p$ ($l_p = \mathrm{argmax}_l a_{\textrm{obj}}^{(l)}$) and the relative increase of $a_{\textrm{obj}}^{(l)}$ in CLVS over vanilla method (clipped at 100\%). The statistical results are shown in \cref{fig:coco}.
In the vanilla method, attention to key objects typically emerges around layer 14 but quickly diminishes, consistent with prior findings~\cite{ViCrop}. In contrast, CLVS sustains focus on key objects, as evidenced by a large relative increase in attention weights in the lower layers, with an average gain of 31.65\% from layers 15 to 17. Since early smoothing termination is applied starting in the middle layers, the model no longer preferentially attends to key objects above these layers. This enhanced attention facilitates better understanding of object-related information, enabling CLVS to reduce attribute and relation hallucinations.

\subsection{Ablation Study}
We conduct an ablation study on AMBER, with results presented in \cref{tab:ablation}.
\paragraph{Unified Position.} We uniform the position ids in visual tokens in the first layer to prevent position bias in visual attention. This method does not show significant effect in image captioning, as the task typically involves longer textual contexts, which dilute the impact of visual position bias compared to visual understanding tasks. Nevertheless, removing this module still leads to a slight increase in hallucination rate. In contrast, for visual understanding tasks, visual position bias can distort the attention distribution over the visual regions, thereby undermining the effectiveness of sustained visual attention.
\paragraph{Early Termination.}
LVLMs primarily process visual information in the early and middle layers; therefore, we terminate CLVS when visual understanding completion to reduce unnecessary computation.
The TTFT results confirm that this mechanism accelerates inference speed while maintaining visual understanding performance with negligible degradation. Moreover, we observe that Early Termination yields particularly notable benefits in image captioning. This is because, when generating tokens with strong language priors, the model relies less on visual input~\cite{CICD} and reaches convergence earlier~\cite{IBD}. In such cases, prolonging visual attention inappropriately can lead to adverse effects and increase hallucinations. Early Termination mitigates this issue by timely deactivating sustained focus, thereby improving both efficiency and generation accuracy.
% We observe positive contributions from both unified positional indices and early termination. For unified positional indices, the improvements are considerably smaller than those achieved by vision smoothing, since this operation is applied only in the first layer and its influence gradually diminishes during subsequent smoothing. Therefore, simply increasing the amount of information attended to by altering positional indices does not directly enhance visual understanding. Regarding early smoothing termination, we confirm that it has no negative impact on model performance, consistent with previous findings that visual understanding in LVLMs primarily occurs in the early and middle layers. Thus, early termination reduces  computational cost without sacrificing performance. As a result, we confirm the effectiveness of all elements in our CLVS.

%\begin{equation}
%    \alpha_{obj}^l = \sum_{i=1}^{N_v} \alpha_i^l \cdot \frac{Area(patch_i \cap obj)}{Area(patch_i)}
%\end{equation}

\paragraph{Effects of $\beta$ and $\gamma$.}

\begin{figure}[t]
  \centering
  \includegraphics[width=0.7\linewidth]{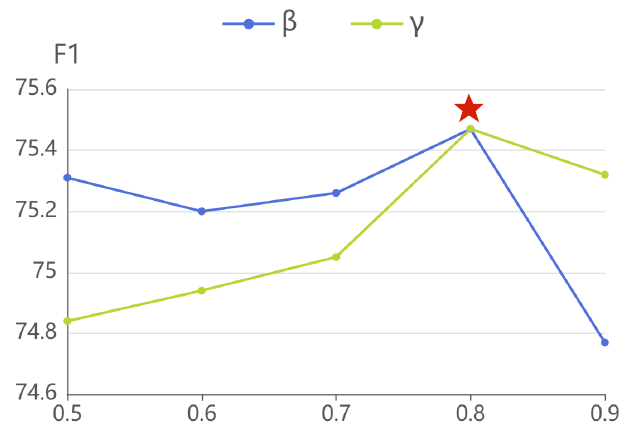}
  \caption{\label{fig:grid}
  % Effectiveness of $\alpha$ and $\beta$.  
  Performance of CLVS with varying values of $\beta$ and $\gamma$. We vary $\beta$ while fixing $\gamma$ at 0.8, and similarly vary $\gamma$ while fixing $\beta$ at 0.8.
  }
\end{figure}

$\beta$ and $\gamma$ are two important hyperparameters that directly affect the performance of our method. We conduct experiments by varying their values, with results shown in \cref{fig:grid}. We find a notable drop in performance when $\beta$ approaches 1 (with $\gamma$ fixed at 0.8), as the vision memory plays a smaller role in smoothing the visual attention distribution. We also observe a performance drop when $\gamma$ is small (with $\beta$ fixed at 0.8), since the vision memory may update too frequently. Very large or very small values of $\beta$ cause dramatic changes in the attention distributions across layers. These results are consistent with our hypothesis that sustained focus on key objects is essential.

\section{Conclusion}
In this paper, we propose Cross-Layer Vision Smoothing (CLVS) to mitigate advantageous attention decay in LVLMs. CLVS initializes a vision memory with unified visual positional indices in the first layer, which is then refined and updated across subsequent layers. Using output uncertainty, we adaptively terminate smoothing once visual perception is complete, thereby reducing redundant computation. Experiments demonstrate that CLVS sustains attention on key objects and enhances visual understanding.

{
    \small
    \bibliographystyle{ieeenat_fullname}
    \bibliography{main}
}

\clearpage
\setcounter{page}{1}
\maketitlesupplementary

\renewcommand\thesection{\Alph{section}}
\setcounter{section}{0}

\begin{table*}
\centering
\setlength{\tabcolsep}{1mm}
% \small{
\caption{\label{tab:mme}
The results on MME. \textbf{Total} denotes the overall score.
}
\begin{tabular}{l|cccccc|cccccc}
\toprule
& \multicolumn{6}{c}{LLaVA-1.5-7B} & \multicolumn{6}{|c}{LLaVA-1.5-13B}\\
\midrule
Method & Exsitence & Count & Position & Color & OCR & Total  & Exsitence & Count & Position & Color & OCR & Total \\

\midrule
Vanilla & 195.0 & 158.3 & 123.3 & 155.0 & 125.0 & 756.7 & 190.0 & 158.3 & 120.0  & 160.0 &  132.5 & 760.8 \\
VCD &  185.0 & 158.3 & 123.3 & 146.7 & 122.5 & 735.8 &  190.0 & 148.3 & 120.0  & 155.0 & 140.0 & 753.3\\
PAI & 185.0 & 151.7 & 128.3 & 170.0 & 132.5 & 767.5 & 190.0 & 153.3 & 105.0 & 165.0 & 117.5 & 730.8 \\
% VAR & 190.0 & 153.3 & 123.3 & 165.0 & 125.0 & 756.7 & \textbf{190.0} & 153.3 & \textbf{120.0}  & 160.0 & 132.5 & 755.8\\
AD-HH & 195.0 & 158.3 & 128.3 & 165.0 & 125.0 & 771.7 & 190.0 & 148.3 & 120.0  & 160.0 & 132.5 & 750.8\\
VAF &  190.0 & 151.7 & 128.3 & 155.0 &  140.0 & 765.0 & 190.0 & 163.3 & 120.0  & 155.0 &  132.5 & 760.8\\
TARAC & 195.0 & 158.3 & 128.3 & 165.0 & 125.0 & 771.7 & 190.0 & 158.3 & 120.0  & 170.0 & 132.5 & 770.8\\
\rowcolor{gray!20}
% CLVS & \textbf{195.0} & \textbf{163.3} & 123.3 & \textbf{170.0} & \textbf{140.0} & \textbf{791.7} & \textbf{190.0} & \textbf{168.3} & 115.0 & \textbf{175.0} &  \textbf{147.5} & \textbf{795.8}\\
CLVS & 195.0 & 158.3 & 123.3 & 175.0 & 132.5 & \textbf{784.1} & 195.0 & 163.3 & 115.0 & 180.0 &  140.0 & \textbf{793.3}\\
\bottomrule
\end{tabular}
% }

\end{table*}

\begin{table*}
 \setlength{\tabcolsep}{3mm}
\centering
\caption{\label{tab:qwen}
Results on AMBER. ``All'' denotes the aggregated results over all test samples.
}
\begin{tabular}{l|l|cc|cc|cc|cc}
\toprule
\multirow{2}{*}{\textbf{Model}} & \multirow{2}{*}{\textbf{Method}} 
& \multicolumn{2}{c|}{\textbf{Existence}} 
& \multicolumn{2}{c|}{\textbf{Attribute}} 
& \multicolumn{2}{c|}{\textbf{Relation}} 
& \multicolumn{2}{c}{\textbf{All}} \\
\cmidrule(lr){3-4} \cmidrule(lr){5-6} \cmidrule(lr){7-8} \cmidrule(lr){9-10}
 & & Acc.  $\uparrow$ & F1 $\uparrow$ & Acc. $\uparrow$ & F1 $\uparrow$ & Acc. $\uparrow$ & F1 $\uparrow$ & Acc. $\uparrow$ & F1 $\uparrow$ \\
\midrule

\multirow{2}{*}{Qwen-Chat} 
  & Vanilla & 95.7 & 97.8 & 81.8 & 83.0 & 55.5 & 64.5 & 83.6 & 88.3
 \\
  & CLVS & \textbf{95.9} & \textbf{97.9} & \textbf{82.2} & \textbf{83.2} & \textbf{57.5} & \textbf{65.1} & \textbf{84.0} & \textbf{88.5}\\
\midrule
\multirow{2}{*}{Qwen2.5-VL} 
  & Vanilla & \textbf{95.3} & \textbf{97.5} & 83.5 & 84.4 & 73.9 & 75.8 & 86.5 & 90.1\\
  & CLVS &  95.0 & 97.4 & \textbf{83.9} & \textbf{84.6} & \textbf{74.5} & \textbf{76.3} & \textbf{86.6} & \textbf{90.2}\\
  
\bottomrule
\end{tabular}
\end{table*}

\section{Detailed Experimental Setup}
To assess the effectiveness of our method in alleviating multimodal hallucination, we conduct evaluations on four widely used benchmarks: one generative benchmark (CHAIR \cite{chair}), two discriminative benchmark (POPE \cite{POPE}, R-Bench \cite{R-Bench}), and one hybrid benchmark (AMBER \cite{AMBER}).

\paragraph{CHAIR.}
CHAIR measures the proportion of hallucinated objects—that is, objects generated by the model but absent from the ground-truth annotations. Following prior work, we randomly sample 500 images from the MSCOCO dataset.
This benchmark reports two metrics, CHAIRs and CHAIRi, defined as:
\begin{equation}
\begin{aligned}
\text{CHAIRs} &= \frac{|\text{Hallucinated Objects}|}{|\text{All Objects}|}, \\
\text{CHAIRi} &= \frac{|\text{Hallucinated Captions}|}{|\text{All Captions}|}.
\end{aligned}
\end{equation}

\paragraph{POPE.}
POPE is a widely used benchmark for measuring object hallucination by prompting LVLMs to judge whether a specific object appears in a given image. It contains three datasets—\emph{MSCOCO}, \emph{A-OKVQA}, and \emph{GQA}—each evaluated under three negative sampling strategies: \emph{Random}, \emph{Popular}, and \emph{Adversarial}.
Each subset consists of 3,000 questions paired with 500 images. Accuracy and F1 score are used as the primary metrics.

\paragraph{AMBER.}
AMBER integrates both generative and discriminative assessments and is conducted on a curated set of 1,004 images. Besides image captioning, the benchmark includes 14,216 questions evaluating hallucinations in object, attribute, and relation understanding.
AMBER provides multiple metrics—\emph{CHAIR}, \emph{Cover}, \emph{Hal}, and \emph{Cog}. Given the annotated object set $A_{obj} = {obj_1^A, obj_2^A, \cdots, obj_n^A}$ and the generated object set $R'{obj}$, these metrics are computed as:
\begin{equation}
\begin{aligned}
\text{CHAIR} &= 1 - \frac{\mathrm{len}(R'{obj} \cap A_{obj})}{\mathrm{len}(R'{obj})},\\
\text{Cover} &= \frac{\mathrm{len}(R'{obj} \cap A_{obj})}{\mathrm{len}(A_{obj})}, \\
\text{Hal} &= \frac{{\text{CHAIR} > 0}}{{\text{All Caps}}}, \\
\text{Cog} &= \frac{\mathrm{len}(R'{obj} \cap H{obj})}{\mathrm{len}(R'{obj})},
\end{aligned}
\end{equation}
where $H{obj}$ denotes the set of hallucinated target objects predicted by the LVLMs, and \emph{All Caps} refers to all generated captions.

\paragraph{R-Bench.}
R-Bench is a benchmark for evaluating relation hallucination, which pose a greater challenge to the model's visual understanding ability.
This benchmark provides multiple randomly constructed subsets with a balanced number of positive and negative samples. We use the subset with ID 1, which contains 5,530 samples.
We conduct experiments using image-level questions.

\subsection{Implementation Details for Baselines}
The hyperparameter settings for the baselines in our experiments are as follows:
\begin{itemize}
    \item \textbf{VCD}~\cite{VCD}: We add 500 steps of Gaussian noise to the original image to construct the contrastive visual input, and the hyperparameters for contrastive decoding are set to $\alpha=1$, $\beta=0.1$.
    \item \textbf{PAI}~\cite{PAI}: We activate the PAI method starting from layer 2 (counting from 0), with hyperparameters set to $\alpha=0.5$, $\gamma=1.1$.
    % \item \textbf{VAR}~\cite{VAR}: The hyperparameters in this method are set as follows: $\tau=20$, $\rho=0.5$, and $p=0.6$.
    \item \textbf{AD-HH}~\cite{AD-HH}: We set the hallucination threshold $\tau$ for attention heads to 0.5. For the definition of hallucination heads, please refer to the original paper.
    \item \textbf{VAF}~\cite{VAF}: We set $\beta=0.1$, $\alpha=0.15$, and activate this method from layer 9 to layer 14 (counting from 0).
    \item \textbf{TARAC}~\cite{TARAC}: We set $\beta=0.5$, $\alpha=0.5$, and activate this method from layer 9 to layer 15 (counting from 0).
    
\end{itemize}
\textbf{Note: Parameters with the same name may have different meanings across methods; please refer to the original papers for their specific definitions.}

\section{Results on MME}
We evaluated the performance of our method on a wider range of tasks using the MME benchmark~\cite{MME}, as shown in Table~\ref{tab:mme}.
Our method achieves the highest overall score among all baselines, but exhibits relatively weak performance on the position perception task.
We argue that the position task relies heavily on spatial information in the image, and our approach—by prolonging LVLMs' focus on key objects—does not enhance their understanding of spatial relationships.
In contrast, our method significantly improves LVLMs' comprehension of key objects in the color, count, and OCR tasks.

\section{Effectiveness on Other MLLMs}
We validate the effectiveness of our method on the Qwen series of models. Since Qwen2.5-VL~\cite{qwen25-vl} employs 3D positional indices, we disable the Unified Position module in our experiments with Qwen2.5-VL. The results in \Cref{tab:qwen} show that CLVS still achieves significant improvements on the Qwen series models, further verifying the positive impact of sustained visual attention in vision understanding.

\end{document}